\documentclass{article} 
\usepackage{iclr2015,times}
\usepackage{hyperref}
\usepackage{url}

\usepackage{natbib}

\usepackage{graphicx}

\usepackage{booktabs}

\usepackage{algorithmic}
\usepackage{algorithm}

\title{Training deep neural networks with \\ low precision multiplications}

\author{
Matthieu Courbariaux \& Jean-Pierre David\\
\'{E}cole Polytechnique de Montr\'{e}al \\
\texttt{\{matthieu.courbariaux,jean-pierre.david\}@polymtl.ca} \\
\And
Yoshua Bengio \\
Universit\'{e} de Montr\'{e}al, CIFAR Senior Fellow\\
\texttt{yoshua.bengio@gmail.com} \\
}

\iclrfinalcopy 

\begin{document}

\maketitle

\begin{abstract}

Multipliers are the most space and power-hungry arithmetic operators of the digital implementation of deep neural networks.
We train a set of state-of-the-art neural networks (Maxout networks) on three benchmark datasets:
MNIST, CIFAR-10 and SVHN. They are trained with three distinct formats: floating point, fixed point and dynamic fixed point.
For each of those datasets and for each of those formats,
we assess the impact of the precision of the multiplications on the final error after training.
We find that {\em very low precision is sufficient}
not just for running trained networks but {\em also for training them}.
For example, it is possible to train Maxout networks with {\bf 10} bits multiplications.\

\end{abstract}

\section{Introduction}

The {\em training} of deep neural networks is very often limited by hardware.
Lots of previous works address the best exploitation of general-purpose hardware,
typically CPU clusters  \citep{Dean-et-al-NIPS2012} and GPUs  \citep{Coates-et-al-2009,Krizhevsky-2012}.
Faster implementations usually lead to state of the art results \citep{Dean-et-al-NIPS2012,Krizhevsky-2012}.

Actually, such approaches always consist in adapting the algorithm to best exploit state of the art general-purpose hardware.
Nevertheless, some dedicated deep learning hardware is appearing as well.
FPGA and ASIC implementations claim a better power efficiency than general-purpose hardware
\citep{Kim-et-al-2009, Farabet-et-al-2011, Pham-et-al-2012, Chen-et-al-ACM2014, Chen-et-al-IEEE2014}.
In contrast with general-purpose hardware, dedicated hardware such as ASIC and FPGA enables to build the hardware from the algorithm.

Hardware is mainly made out of memories and arithmetic operators.
Multipliers are the most space and power-hungry arithmetic operators of the digital implementation of deep neural networks.
The objective of this article is to assess the possibility to
reduce the precision of the multipliers for deep learning:
\begin{itemize}

  \item We train deep neural networks with low precision multipliers and high precision accumulators (Section \ref{sec:accumulations}).

  \item We carry out experiments with three distinct formats:
    \begin{enumerate}
      \item Floating point (Section \ref{sec:float})
      \item Fixed point (Section \ref{sec:fixed})
      \item Dynamic fixed point, which we think is a good compromise between floating and fixed points
        (Section \ref{sec:dynamic})
    \end{enumerate}

  \item We use a higher precision for the parameters during the updates
    than during the forward and backward propagations (Section \ref{sec:update}).

  \item Maxout networks  \citep{Goodfeli-et-al-TR2013} are a set of state-of-the-art neural networks (Section \ref{sec:maxout}).
    We train Maxout networks with slightly less capacity than \citet{Goodfeli-et-al-TR2013}
    on three benchmark datasets: MNIST, CIFAR-10 and SVHN (Section \ref{sec:baseline}).

  \item For each of the three datasets and for each of the three formats,
    we assess the impact of the precision of the multiplications on the final error of the training.
    We find that {\em very low precision multiplications are sufficient} not just for running trained networks but
    {\em also for training them} (Section \ref{sec:low}).
    We made our code available
    \footnote{ \url{https://github.com/MatthieuCourbariaux/deep-learning-multipliers} }.

\end{itemize}

\section{Multiplier-accumulators}
\label{sec:accumulations}

\begin{table}[h]
\center
\begin{tabular}{@{}ccc@{}}
\toprule
Multiplier (bits) & Accumulator (bits) & Adaptive Logic Modules (ALMs) \\ \midrule
32                & 32                 & 504                           \\
16                & 32                 & 138                           \\
16                & 16                 & 128                           \\ \bottomrule
\end{tabular}
\caption{Cost of a fixed point multiplier-accumulator on a Stratix V Altera FPGA.}
\label{tab:mac}%
\end{table}%

\begin{algorithm}[h]
\begin{algorithmic}
    \FORALL{layers}
    \STATE Reduce the precision of the parameters and the inputs
    \STATE Apply convolution or dot product (with high precision accumulations)
    \STATE Reduce the precision of the weighted sums
    \STATE Apply activation functions
    \ENDFOR
    \STATE Reduce the precision of the outputs
\end{algorithmic}
\caption{Forward propagation with low precision multipliers.}
\label{alg:prop}
\end{algorithm}

Applying a deep neural network (DNN) mainly consists in convolutions and matrix multiplications.
The key arithmetic operation of DNNs is thus the multiply-accumulate operation.
Artificial neurons are basically multiplier-accumulators computing weighted sums of their inputs.

The cost of a fixed point multiplier varies as the square of the precision (of its operands) for small widths while the cost of adders and accumulators varies as a linear function of the precision \citep{4302704}.
As a result, the cost of a fixed point multiplier-accumulator mainly depends on the precision of the multiplier, 
as shown in table \ref{tab:mac}.
In modern FPGAs, the multiplications can also be implemented with dedicated DSP blocks/slices. One DSP block/slice can implement a single $27 \times 27$ multiplier, a double $18 \times 18$ multiplier or a triple $9 \times 9$ multiplier. Reducing the precision can thus lead to a gain of 3 in the number of available multipliers inside a modern FPGA.

In this article, we train deep neural networks with low precision multipliers and high precision accumulators,
as illustrated in Algorithm \ref{alg:prop}.

\section{Floating point}
\label{sec:float}

\begin{figure}[ht]
\begin{center}
\centerline{\includegraphics[width=.75\textwidth]{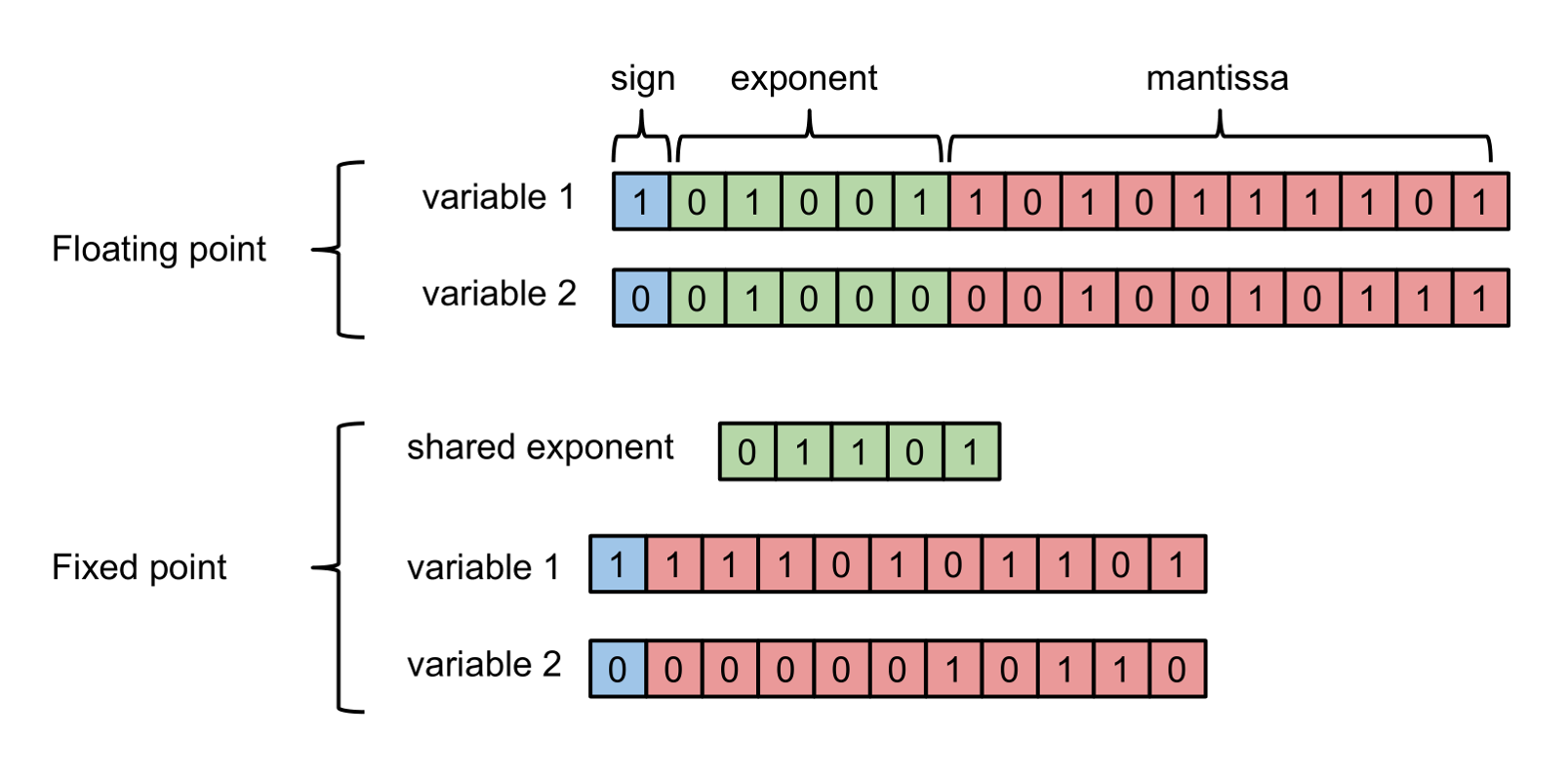}}
\end{center}
\caption{Comparison of the floating point and fixed point formats.}
\label{fig:exponent}
\end{figure}

\begin{table}[h]
\begin{center}
\begin{tabular}{@{}llll@{}}
\toprule
Format                          &  Total bit-width & Exponent bit-width & Mantissa bit-width \\ \midrule
Double precision floating point &  64              & 11                 & 52                 \\
Single precision floating point &  32              & 8                  & 23                 \\
Half precision floating point   &  16              & 5                  & 10                 \\ \bottomrule
\end{tabular}
\end{center}
\caption{Definitions of double, single and half precision floating point formats.
}
\label{tab:float_definitions}
\end{table}

Floating point formats are often used to represent real values.
They consist in a sign, an exponent, and a mantissa, as illustrated in figure \ref{fig:exponent}.
The exponent gives the floating point formats a wide range, and the mantissa gives them a good precision.
One can compute the value of a single floating point number using the following formula:
\[
  value = (-1)^{sign} \times \left(1 + \frac{mantissa}{2^{23}}\right) \times 2^{(exponent-127)}
\]

Table \ref{tab:float_definitions} shows the exponent and mantissa widths associated with each floating point format.
In our experiments, we use single precision floating point format as our reference
because it is the most widely used format in deep learning, especially for GPU computation.
We show that the use of half precision floating point format has little to no impact on the training of neural networks.
At the time of writing this article, no standard exists below the half precision floating point format.

\section{Fixed point}
\label{sec:fixed}

Fixed point formats consist in a signed mantissa and a global scaling factor shared between all fixed point variables.
The scaling factor can be seen as the position of the radix point.
It is usually fixed, hence the name "fixed point".
Reducing the scaling factor reduces the range and augments the precision of the format.
The scaling factor is typically a power of two for computational efficiency (the scaling multiplications are replaced with shifts).
As a result, fixed point format can also be seen as a floating point format with a {\em unique shared fixed exponent}
, as illustrated in figure \ref{fig:exponent}.
Fixed point format is commonly found on embedded systems with no FPU (Floating Point Unit).
It relies on integer operations. It is hardware-wise cheaper than its floating point counterpart,
as the exponent is shared and fixed.

\section{Dynamic fixed point}
\label{sec:dynamic}

\begin{algorithm}[h]
\begin{algorithmic}
    \REQUIRE a matrix $M$, a scaling factor $s_{t}$,
        and a maximum overflow rate $r_{max}$.
    \ENSURE an updated scaling factor $s_{t+1}$.
    \IF{the overflow rate of $M>r_{max}$}
        \STATE $s_{t+1} \leftarrow 2 \times s_{t}$
    \ELSIF{the overflow rate of $2 \times M \le r_{max}$}
        \STATE $s_{t+1} \leftarrow s_{t} /2$
    \ELSE
        \STATE $s_{t+1} \leftarrow s_{t}$
    \ENDIF
\end{algorithmic}
\caption{Policy to update a scaling factor.}
\label{alg:DFXP}
\end{algorithm}

When training deep neural networks,
\begin{enumerate}
  \item activations, gradients and parameters have {\em very different ranges}.
  \item gradients ranges {\em slowly diminish} during the training.
\end{enumerate}
As a result, the fixed point format, with its unique shared fixed exponent, is ill-suited to deep learning.

The dynamic fixed point format \citep{Williamson-1991}
is a variant of the fixed point format in which there are {\em several scaling factors} instead of a single global one.
Those scaling factors are {\em not fixed}. As such, it can be seen as a compromise between floating point format
- where each scalar variable owns its scaling factor which is updated during each operations -
and fixed point format - where there is only one global scaling factor which is never updated.
With dynamic fixed point, a few grouped variables share a scaling factor which is updated from time to time
to reflect the statistics of values in the group.

In practice, we associate each layer's weights, bias, weighted sum, outputs (post-nonlinearity)
and the respective gradients vectors and matrices with a different scaling factor.
Those scaling factors are initialized with a global value.
The initial values can also be found during the training with a higher precision format.
During the training, we update those scaling factors at a given frequency,
following the policy described in Algorithm \ref{alg:DFXP}.

\section{Updates vs. propagations}
\label{sec:update}

We use a higher precision for the parameters during the updates than during the forward and backward propagations,
respectively called fprop and bprop.
The idea behind this is to be able to accumulate small changes in the parameters (which
requires more precision) and while on the other hand sparing a few bits of memory bandwidth
during fprop. This can
be done because of the implicit averaging performed via stochastic gradient descent during training:
\[
  \theta_{t+1} = \theta_t - \epsilon \frac{\partial C_t(\theta_t)}{\partial \theta_t}
\]
where $C_t(\theta_t)$ is the cost to minimize over the minibatch visited at iteration
$t$ using $\theta_t$ as parameters
and $\epsilon$ is the learning rate. We see that the resulting parameter is
the sum
\[
  \theta_T = \theta_0 - \epsilon \sum_{t=1}^{T-1} \frac{\partial C_t(\theta_t)}{\partial \theta_t}.
\]
The terms of this sum are not statistically independent (because the value of $\theta_t$ depends
on the value of $\theta_{t-1}$) but the dominant variations come from the random sample
of examples in the minibatch ($\theta$ moves slowly) so that a strong averaging
effect takes place, and each contribution in the sum is relatively small, hence
the demand for sufficient precision (when adding a small number with a large number).

\section{Maxout networks}
\label{sec:maxout}

A Maxout network is a multi-layer neural network that uses maxout units in its hidden layers.
A maxout unit outputs the maximum of a set of $k$ dot products between $k$ weight
vectors and the input vector of the unit (e.g., the output of the previous layer):
\[
  h^l_i = \max_{j=1}^k ( b^l_{i,j} + w^l_{i,j} \cdot h^{l-1} )
\]
where $h^l$ is the vector of activations at layer $l$ and weight vectors $w^l_{i,j}$
and biases $b^l_{i,j}$ are the parameters of the $j$-th filter of unit $i$ on layer $l$.

A maxout unit can be seen as a generalization of the rectifying units
 \citep{Jarrett-ICCV2009,Nair-2010-small,Glorot+al-AI-2011-small,Krizhevsky-2012-small}
\[
  h^l_i = \max(0, b^l_i + w^l_i \cdot h^{l-1})
\]
which corresponds to a maxout unit when $k=2$ and one of the filters is forced at 0 \citep{Goodfeli-et-al-TR2013}.
Combined with dropout, a very effective regularization method  \citep{Hinton-et-al-arxiv2012},
maxout networks achieved state-of-the-art results on a number of benchmarks  \citep{Goodfeli-et-al-TR2013},
both as part of fully connected feedforward deep nets and as part of deep convolutional nets.
The dropout technique provides a good approximation of model averaging with shared parameters
across an exponentially large number of networks that are formed by subsets of the units
of the original noise-free deep network.

\section{Baseline results}
\label{sec:baseline}

We train Maxout networks with slightly less capacity than \citet{Goodfeli-et-al-TR2013}
on three benchmark datasets: MNIST, CIFAR-10 and SVHN.
In Section \ref{sec:low}, we use the same hyperparameters as in this section
to train Maxout networks with low precision multiplications.

\begin{table}[h]
\begin{center}
\begin{tabular}{@{}lllll@{}}
\toprule
Dataset   & Dimension                       & Labels           & Training set   & Test set  \\ \midrule
MNIST      & 784 (28 $\times$ 28 grayscale)  & 10               & 60K            & 10K       \\
CIFAR-10    & 3072 (32 $\times$ 32 color)     & 10               & 50K            & 10K       \\
SVHN       & 3072 (32 $\times$ 32 color)     & 10               & 604K           & 26K       \\ \bottomrule
\end{tabular}
\end{center}
\caption{
Overview of the datasets used in this paper.
}
\label{tab:datasets}
\end{table}

\begin{table}[h]
\begin{center}
\begin{tabular}{@{}lllllll@{}}
\toprule
Format                          & Prop. & Up. & PI MNIST   & MNIST  & CIFAR-10 & SVHN   \\ \midrule
 \citet{Goodfeli-et-al-TR2013}  & 32    & 32  & 0.94\%     & 0.45\% & 11.68\% & 2.47\% \\
Single precision floating point & 32    & 32  & 1.05\%     & 0.51\% & 14.05\% & 2.71\% \\
Half precision floating point   & 16    & 16  & 1.10\%     & 0.51\% & 14.14\% & 3.02\% \\
Fixed point                     & 20    & 20  & 1.39\%     & 0.57\% & 15.98\% & 2.97\% \\
Dynamic fixed point             & 10    & 12  & 1.28\%     & 0.59\% & 14.82\% & 4.95\% \\ \bottomrule
\end{tabular}
\end{center}
\caption{
Test set error rates of single and half floating point formats, fixed and dynamic fixed point formats on
the permutation invariant (PI) MNIST, MNIST (with convolutions, no distortions),
CIFAR-10 and SVHN datasets.
{\bf Prop.} is the bit-width of the propagations and {\bf Up.}
is the bit-width of the parameters updates.
The single precision floating point line refers to the results of our experiments.
It serves as a baseline to evaluate the degradation brought by lower precision.
}
\label{tab:all_results}
\end{table}

\subsection{MNIST}

The MNIST \citep{LeCun+98} dataset is described in Table \ref{tab:datasets}.
We do not use any data-augmentation (e.g. distortions) nor any unsupervised pre-training.
We simply use minibatch stochastic gradient descent (SGD) with momentum.
We use a linearly decaying learning rate and a linearly saturating momentum.
We regularize the model with dropout and a constraint on the norm of each weight vector, as in \citep{Srebro05}.

We train two different models on MNIST.
The first is a permutation invariant (PI) model which is unaware of the structure of the data.
It consists in two fully connected maxout layers followed by a softmax layer.
The second model consists in three convolutional maxout hidden layers (with spatial max pooling on top of the maxout layers)
followed by a densely connected softmax layer.

This is the same procedure as in \citet{Goodfeli-et-al-TR2013},
except that we do not train our model on the validation examples.
As a consequence, our test error is slightly larger than the one reported in \citet{Goodfeli-et-al-TR2013}.
The final test error is in Table \ref{tab:all_results}.

\subsection{CIFAR-10}

Some comparative
characteristics of the CIFAR-10 \citep{KrizhevskyHinton2009} dataset are given in Table \ref{tab:datasets}.
We preprocess the data using global contrast normalization and ZCA whitening.
The model consists in three convolutional maxout layers, a fully connected maxout layer, and a fully connected softmax layer.
We follow a similar procedure as with the MNIST dataset.
This is the same procedure as in \citet{Goodfeli-et-al-TR2013},
except that we reduced the number of hidden units and
that we do not train our model on the validation examples.
As a consequence, our test error is slightly larger than the one reported in \citet{Goodfeli-et-al-TR2013}.
The final test error is in Table \ref{tab:all_results}.

\subsection{Street View House Numbers}

The SVHN \citep{Netzer-wkshp-2011} dataset is described in Table \ref{tab:datasets}.
We applied local contrast normalization preprocessing the same way as \citet{Zeiler+et+al-ICLR2013}.
The model consists in three convolutional maxout layers, a fully connected maxout layer, and a fully connected softmax layer.
Otherwise, we followed the same approach as on the MNIST dataset.
This is the same procedure as in \citet{Goodfeli-et-al-TR2013},
except that we reduced the length of the training.
As a consequence, our test error is bigger than the one reported in \citet{Goodfeli-et-al-TR2013}.
The final test error is in Table \ref{tab:all_results}.

\section{Low precision results}
\label{sec:low}

\begin{figure}[ht]
\begin{center}
\centerline{\includegraphics[width=.66\textwidth]{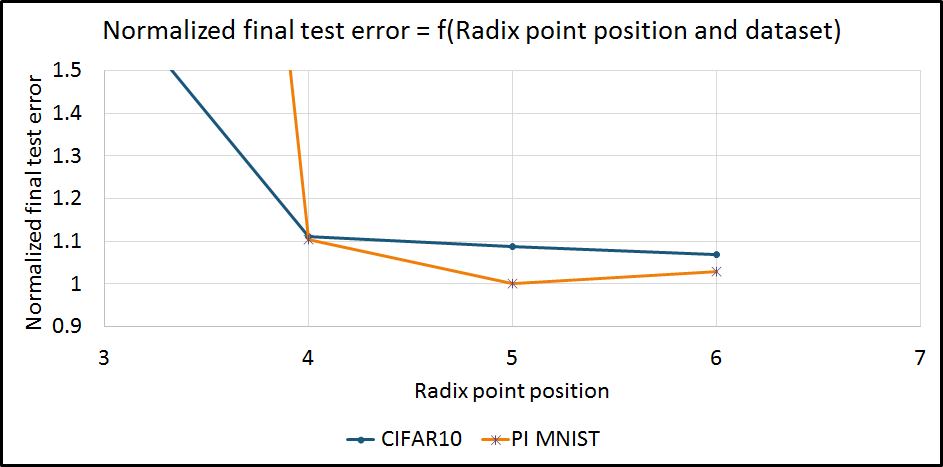}}
\end{center}
\caption{
Final test error depending on the radix point position (5 means after the $5$th most significant bit)
and the dataset (permutation invariant MNIST and CIFAR-10).
The final test errors are normalized, that is to say divided by the dataset single float test error.
The propagations and parameter updates bit-widths are both set to 31 bits (32 with the sign).
}
\label{fig:range}
\end{figure}

\begin{figure}[ht]
\begin{center}
\centerline{\includegraphics[width=\textwidth]{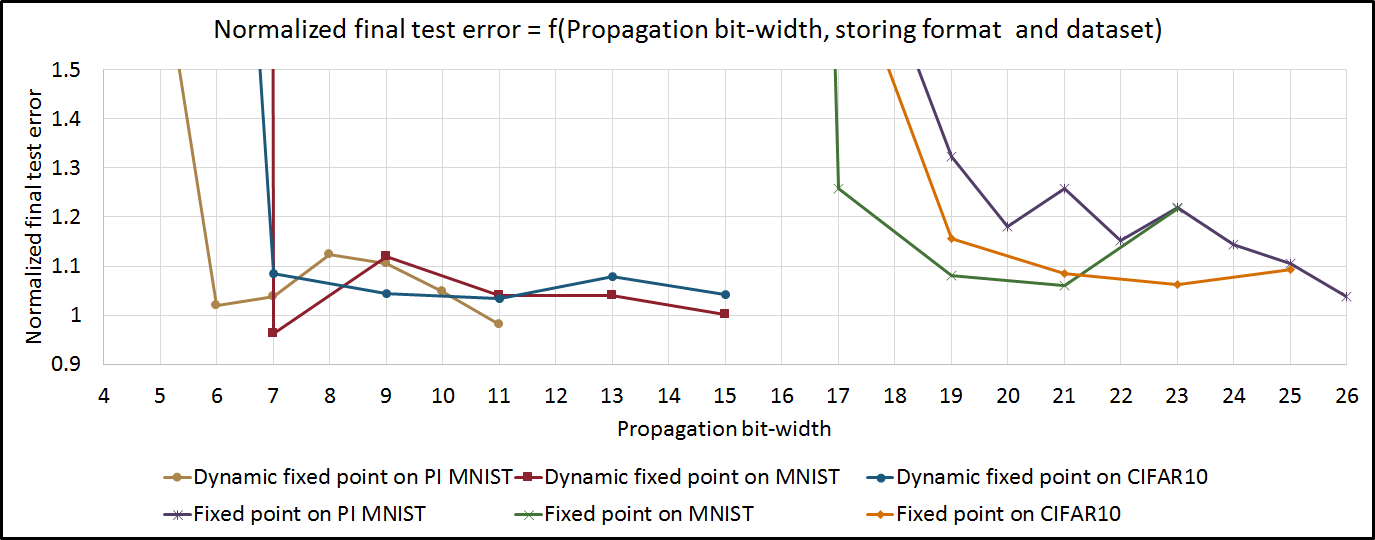}}
\end{center}
\caption{
Final test error depending on the propagations bit-width, the format (dynamic fixed or fixed point)
and the dataset (permutation invariant MNIST, MNIST and CIFAR-10).
The final test errors are normalized, which means that they are divided by the dataset single float test error.
For both formats, the parameter updates bit-width is set to 31 bits (32 with the sign).
For fixed point format, the radix point is set after the fifth bit.
For dynamic fixed point format, the maximum overflow rate is set to 0.01\%.
}
\label{fig:propagations}
\end{figure}

\begin{figure}[ht]
\begin{center}
\centerline{\includegraphics[width=\textwidth]{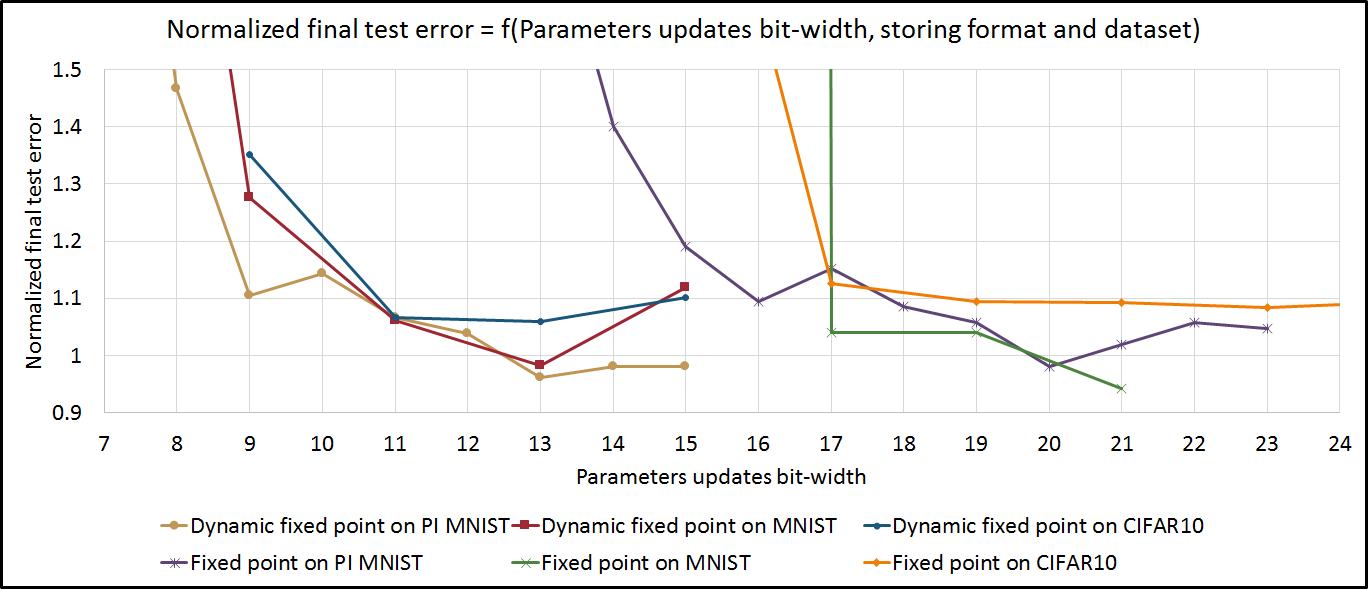}}
\end{center}
\caption{
Final test error depending on the parameter updates bit-width, the format (dynamic fixed or fixed point)
and the dataset (permutation invariant MNIST, MNIST and CIFAR-10).
The final test errors are normalized, which means that they are divided by the dataset single float test error.
For both formats, the propagations bit-width is set to 31 bits (32 with the sign).
For fixed point format, the radix point is set after the fifth bit.
For dynamic fixed point format, the maximum overflow rate is set to 0.01\%.
}
\label{fig:updates}
\end{figure}

\begin{figure}[ht]
\begin{center}
\centerline{\includegraphics[width=.75\textwidth]{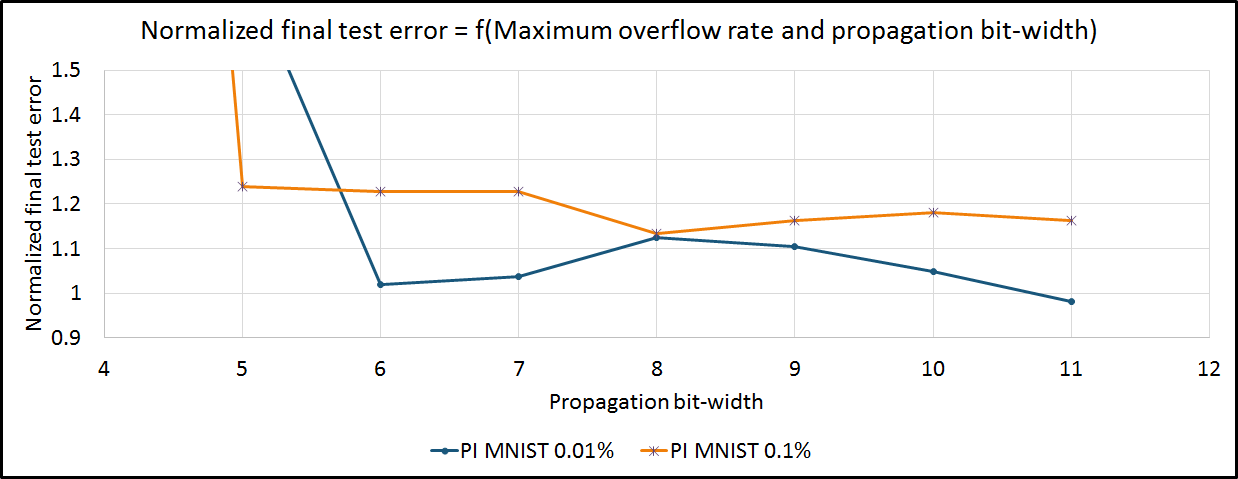}}
\end{center}
\caption{
Final test error depending on the maximum overflow rate and the propagations bit-width.
The final test errors are normalized, which means that they are divided by the dataset single float test error.
The parameter updates bit-width is set to 31 bits (32 with the sign).
}
\label{fig:overflow}
\end{figure}

\subsection{Floating point}

Half precision floating point format has little to no impact on the test set error rate,
as shown in Table \ref{tab:all_results}.
We conjecture that a high-precision fine-tuning could recover the small degradation of the error rate.

\subsection{Fixed point}

The optimal radix point position in fixed point is after the fifth (or arguably the sixth) most important bit,
as illustrated in Figure \ref{fig:range}.
The corresponding range is approximately [-32,32].
The corresponding scaling factor depends on the bit-width we are using.
The minimum bit-width for propagations in fixed point is 19 (20 with the sign).
Below this bit-width, the test set error rate rises very sharply,
as illustrated in Figure \ref{fig:propagations}.
The minimum bit-width for parameter updates in fixed point is 19 (20 with the sign).
Below this bit-width, the test set error rate rises very sharply,
as illustrated in Figure \ref{fig:updates}.
Doubling the number of hidden units does not allow any further reduction of the bit-widths on the permutation invariant MNIST.
In the end, using 19 (20 with the sign) bits for both the propagations and the parameter updates
has little impact on the final test error,
as shown in Table \ref{tab:all_results}.

\subsection{Dynamic fixed point}

We find the initial scaling factors by training with a higher precision format.
Once those scaling factors are found, we reinitialize the model parameters.
We update the scaling factors once every 10000 examples.
Augmenting the maximum overflow rate allows us to reduce the propagations bit-width
but it also significantly augments the final test error rate,
as illustrated in Figure \ref{fig:overflow}.
As a consequence, we use a low maximum overflow rate of 0.01\% for the rest of the experiments.
The minimum bit-width for the propagations in dynamic fixed point is 9 (10 with the sign).
Below this bit-width, the test set error rate rises very sharply,
as illustrated in Figure \ref{fig:propagations}.
The minimum bit-width for the parameter updates in dynamic fixed point is 11 (12 with the sign).
Below this bit-width, the test set error rate rises very sharply,
as illustrated in Figure \ref{fig:updates}.
Doubling the number of hidden units does not allow any further reduction of the bit-widths on the permutation invariant MNIST.
In the end, using 9 (10 with the sign) bits for the propagations and 11 (12 with the sign) bits for the parameter updates
has little impact on the final test error, with the exception of the SVHN dataset,
as shown in Table \ref{tab:all_results}.
This is significantly better than fixed point format, which is consistent with our predictions of Section \ref{sec:dynamic}.

\section{Related works}

\citet{Vanhoucke-et-al-2011} use 8 bits linear quantization to store activations and weights.
Weights are scaled by taking their maximum magnitude in each layer and normalizing them to fall in the [-128, 127] range.
The total memory footprint of the network is reduced by between 3$\times$ and 4$\times$.
This is very similar to the dynamic fixed point format we use (Section \ref{sec:dynamic}).
However, \citet{Vanhoucke-et-al-2011} only {\em apply} already trained neural networks while we actually {\em train} them.

Training neural networks with low precision arithmetic has already been done in previous works
\citep{Holt-et-al-1991,Presley-et-al-1994,Simard+Graf-NIPS1994,Wawrzynek-et-al-IEEE1996,Savich-et-al-2007}
\footnote{
A very recent work \citep{Gupta-et-al-2015} also trains neural networks with low precision.
The authors propose to replace round-to-nearest with stochastic rounding,
which allows to reduce the numerical precision to 16 bits while using the fixed point format.
It would be very interesting to combine dynamic fixed point and stochastic rounding.
}.
Our work is nevertheless original in several regards:
\begin{itemize}
    \item We are the first to train deep neural networks with the dynamic fixed point format.
    \item We use a higher precision for the weights during the updates.
    \item We train some of the latest models on some of the latest benchmarks.
\end{itemize}

\section{Conclusion and future works}

We have shown that:
\begin{itemize}
  \item Very low precision multipliers are sufficient for training deep neural networks.
  \item Dynamic fixed point seems well suited for training deep neural networks.
  \item Using a higher precision for the parameters during the updates helps.
\end{itemize}

Our work can be exploited to:
\begin{itemize}
    \item Optimize memory usage on general-purpose hardware \citep{Gray-et-al-2015}.
    \item Design very power-efficient hardware dedicated to deep learning.
\end{itemize}

There is plenty of room for extending our work:
\begin{itemize}
  \item Other tasks than image classification.
  \item Other models than Maxout networks.
  \item Other formats than floating point, fixed point and dynamic fixed point.
\end{itemize}

\section{Acknowledgement}

We thank the developers of Theano \citep{bergstra+al:2010-scipy,Bastien-Theano-2012},
a Python library which allowed us to easily develop a fast and optimized code for GPU.
We also thank the developers of Pylearn2 \citep{pylearn2_arxiv_2013},
a Python library built on the top of Theano which allowed us to easily interface the datasets with our Theano code.
We are also grateful for funding from NSERC, the Canada Research Chairs, Compute Canada, and CIFAR.

\bibliography{strings,strings-shorter,aigaion,ml,precision}
\bibliographystyle{natbib}

\end{document}